**6**

# EM-Based Mixture Models Applied to Video Event Detection

Alessandra Martins Coelho[1] and Vania Vieira Estrela[2]
[1]*Instituto Federal de Educacao, Ciencia e Tecnologia do Sudeste de Minas Gerais (IF SUDESTE MG), Rio Pomba, MG,*
[2]*Departamento de Telecomunicacoes, Universidade Federal Fluminense (UFF), Niterói, RJ, Brazil*

## 1. Introduction

Surveillance system (SS) development requires hi-tech support to prevail over the shortcomings related to the massive quantity of visual information from SSs [Fuentes & Velastin (2001)]. Anything but reduced human monitoring became impossible by means of its physical and economic implications, and an advance towards an automated surveillance becomes the only way out.

When it comes to a computer vision system, automatic video event comprehension is a challenging task due to motion clutter, event understanding under complex scenes, multi-level semantic event inference, contextualization of events and views obtained from multiple cameras, unevenness of motion scales, shape changes, occlusions and object interactions among lots of other impairments. In recent years, state-of-the-art models for video event classification and recognition [Zhang et al. (2011), Yacoob et al. (1999)] include modeling events to discern context, detecting incidents with only one camera (Ma et al. (2009), Zhao et al. (2002), Zelnik-Manor et al. (2006)], low-level feature extraction and description, high-level semantic event classification and recognition. Even so, it is still very burdensome to recuperate or label a specific video part relying solely on its content.

Principal component analysis (PCA) has been widely known and used, but when combined with other techniques such as the expectation-maximization (EM) algorithm its computation becomes more efficient.

This chapter introduces advances associated to the concept of Probabilistic PCA (PPCA) analysis [Tipping et al., 1999)] by of video event understanding technologies. The PPCA model-based method results from the combination of a linear model and the EM algorithm in an iterative fashion in order to determine a principal subspace (PS). Thus, additional work may be needed to find precise principal eigenvectors of the data covariance matrix, with no rotational uncertainty.

Kernel principal component analysis (KPCA) is a nonlinear PCA extension that relies on the kernel trick. It has received immense consideration for its value in nonlinear feature mining and other applications. On the other hand, the main drawback of the standard KPCA is that the huge amount of computation required, and the space needed to store the kernel matrix. KPCA can be viewed as a primal space problem with samples created via incomplete Cholesky decomposition. Therefore, all the efficient PCA algorithms can be easily adapted into KPCA. Furthermore, KPCA can be extended to a mixture of local KPCA models by applying the mixture model to probabilistic PCA in the primal space. The theoretical analysis and experiments can shed light on the performance of the proposed methods in terms of computational efficiency and storage space, as well as recognition rate, especially when the number of data points $n$ is large.

By considering KPCA from a probabilistic point of view with the help of the EM algorithm, the computational load can be alleviated, but there still exists a rotational ambiguity with the resulting algorithm implementation. To unravel this intricacy, a constrained EM algorithm for KPCA (and PCA) was formulated founded on a coupled probability model. This brings in advantages related to many factors such as the necessary precision of extracted components, the number of the separated smaller data sets (which is usually empirically set), and the data to be processed. As a generic methodology, another thread of speeding up kernel machine learning is to seek a low-rank approximation to the kernel matrix. Since, as noted by several researchers, the spectrum of the kernel matrix tends to decay rapidly, the low-rank approximation often achieves satisfactory precision.

This chapter also aims at looking closely to ways and metrics in order to evaluate these less intensive EM implementations of PCA and KPCA.

## 2. Different ways of computing PCA

PCA is based on statistical properties of vector representations. It is an important tool for image processing because it decorrelates the data and compacts information [Xu (1998), Rosipal & Girolami (2001)].

PCA has been used profusely in all forms of analysis, since it is a straightforward, nonparametric way of extracting important information from ambiguous data sets. It helps reducing an intricate data set to a lower dimensional one that too often expose an unknown and simplified structure.

This section introduces three ways of calculating principal components (PCs): (i) via explicit computation of the covariance matrix $C_X$ or, equivalently, $XX^T$; (ii) by means of the singular value decomposition (SVD) of the original problem so that it can be replaced by the calculation of $C_X=Y^TY$ with $Y=(n-1)^{1/2}X^T$, which requires the determination of eigenvectors of a system with smaller dimension; and (iii) using the EM algorithm.

Because (i) and (ii) are related to the concept of covariance, they require square matrices ($XX^T$ and $Y^TY$, respectively). In the third case, there is an underlying probabilistic interpretation of the problem.

### 2.1 First approach: Solving PCA using eigenvectors of the covariance matrix

By seeking another basis, which is a linear combination of the original basis, the data set can be better represented. Linearity simplifies the problem because it restricts the set of prospective bases, and handles the implicit postulation of continuity in a data set. Let *X* be a matrix representing the original data set *Y*, be another matrix related by a linear transformation *P* that represents a change of basis. *X* is the original recorded data set and *Y* is its new representation

$$PX = Y. \tag{1}$$

Geometrically, *P* is a rotation and a stretch which again transforms *X* into *Y*. The covariance matrix of *X* is

$$C_X = \frac{1}{(n-1)} XX^T \tag{2}$$

PCA also takes for granted that mean and variance are sufficient statistics are enough to depict a probability distribution. This happens to be the case with exponential distributions (Gaussian, Exponential, etc). Deviations from an exponential distribution could nullify this assumption. Diagonalizing a covariance matrix might not give acceptable results. This hypothesis guarantees that the *SNR* and the covariance matrix totally portray noise and redundancies. The following factors can corrupt data: noise, rotation and redundancy. A common noise metric is the signal-to-noise ratio (*SNR*), or a ratio of variances $\sigma^2$ as follows:

$$SNR = \frac{\sigma_{signal}^2}{\sigma_{noise}^2} \tag{3}$$

A high $SNR (\gg 1)$ indicates clean data, while a low *SNR* points to noisy data. Large variances have important dynamics which means that the data is supposed to have a high *SNR*. Thus, principal components (PCs) with larger associated variances correspond to interesting dynamics, while those with lower variances may characterize noise.

Returning to (1), *X is* an *m*×*n* matrix, where *m* is the number of measurement types and *n* is the number of samples. The goal is to find an orthonormal *P* such that

$$C_Y = \frac{1}{(n-1)} YY^T \tag{4}$$

is diagonal and rows of *P* are the *PCs* of *X*. Because lots of real world data are normally distributed, PCA usually provides a robust solution to small deviations from this assumption. Rewriting $C_Y$ in terms of *P* yields

$$\begin{aligned} C_Y &= \frac{1}{(n-1)} YY^T = \frac{1}{(n-1)} (PX)(PX)^T \\ &= \frac{1}{(n-1)} PXX^T P^T = \frac{1}{(n-1)} P(XX^T) P^T \Rightarrow \\ C_Y &= \frac{1}{(n-1)} PAP^T \end{aligned} \tag{5}$$

where $A \equiv XX^T$ is symmetric with

$$A = EDE^T. \tag{6}$$

The eigenvectors of *A* are arranged as columns of *E* and ***D*** is a diagonal matrix. *A* has $r \leq m$ orthonormal eigenvectors where $r$ is the rank of the matrix. The rank of *A* is less than $m$ when *A* is degenerate or all data reside in a subspace of dimension $r \leq m$. Maintaining the constraint of orthogonality, this situation can be remediated by selecting $(m-r)$ further orthonormal vectors to complete *E*. These additional vectors do not influence the final solution because the variances associated with these directions are zero.

Each row **p**$_i$ is an eigenvector of $XX^T$ and form $\boldsymbol{P} \equiv \boldsymbol{E}^T$. Combining the previous equations results in

$$A = \boldsymbol{P}^T DP. \tag{7}$$

Since $P^{-1} = P^T$, then $C_Y$ becomes

$$\begin{aligned}
C_Y &= \frac{1}{(n-1)} \mathbf{P} A \mathbf{P}^T = \frac{1}{(n-1)} \mathbf{P}(\mathbf{P}^T \mathbf{D} \mathbf{P}) \mathbf{P}^T \\
&= \frac{1}{(n-1)} (\mathbf{P}\mathbf{P}^T) \mathbf{D} (\mathbf{P}\mathbf{P}^T) = \frac{1}{(n-1)} (\mathbf{P}\mathbf{P}^{-1}) \mathbf{D} (\mathbf{P}\mathbf{P}^{-1}) \\
&= \frac{1}{(n-1)} \mathbf{D}
\end{aligned} \tag{8}$$

In practice, computing *PCA* of a data set *X* requires subtracting off the mean of each measurement type and the calculation of the eigenvectors of $\boldsymbol{XX^T}$.

## 2.2 A more general solution: SVD

PCA relates closely to singular value decomposition (*SVD*), but *SVD* is a more general method to deal with change of basis. Let *X* be an arbitrary $n \times m$ matrix and $\boldsymbol{X^TX}$ be a symmetric square $n \times n$ matrix with rank *r*. $V = \{v_1, v_2, \ldots, v_r, 0, \ldots, 0\}$ is the set of orthonormal eigenvectors associated with eigenvalues $\boldsymbol{\Sigma} = Diag\{\sigma_1, \sigma_2, \ldots, \sigma_r, 0, \ldots, 0\}$ for the symmetric matrix $X^TX$ such that

$$(\boldsymbol{X^T X})\boldsymbol{v}_i = \lambda_i \boldsymbol{v}_i, \tag{9}$$

where $\sigma_i \equiv \sqrt{\lambda_i}$ are positive real singular values and $U = \{u_1, u_2, \ldots, u_r, 0, \ldots, 0\}$ is the set of orthonormal vectors defined by $u_i = (1/\sigma_i) X v_i$. *V* and *U* contain, respectively, $(m-r)$ and $(n-r)$ appended zeros and $\sigma_1 \geq \sigma_2 \geq \cdots \geq \sigma_r$ are the rank-ordered set of singular values. The matrix version of *SVD* is given by

$$XV = U\Sigma. \tag{10}$$

Because *V* is orthogonal, multiplying both sides of the expression above by $\boldsymbol{V^{-1}} = \boldsymbol{V^T}$ leads to the final form of the SVD:

$$\boldsymbol{X = U\Sigma V^T}, \tag{11}$$

which states that any arbitrary matrix *X* can be converted to an orthogonal matrix, a diagonal matrix and another orthogonal matrix as follows:

$$\mathbf{X} = \mathbf{U\Sigma V}^T \Rightarrow \mathbf{U}^T \mathbf{X} = \mathbf{\Sigma V}^T \Rightarrow \mathbf{U}^T \mathbf{X} = \mathbf{Z}, \tag{12}$$

where $Z \equiv \Sigma V^T$. Note that $U^T$ is a change of basis from $X$ to $Z$. The fact that the orthonormal basis $U^T$ transforms column vectors means that $U^T$ is a basis that spans the columns of $X$. Bases that span the columns are termed the column space of $X$. If $Z \equiv U^T \Sigma$, then the rows of $V^T$ (or the columns of $V$) are an orthonormal basis for transforming $X^T$ into $Z$. Because of the transpose of $X$, it follows that $V$ is an orthonormal basis spanning the row space of $X$.

Matrices $V$ and $U$ are $m \times m$ and $n \times n$ respectively. $\Sigma$ is a matrix with a small amount of non-zero values along its diagonal. The SVD allows for creating a new $m \times n$ matrix $Y$ as follows:

$$Y \equiv \frac{1}{\sqrt{n-1}} X^T, \qquad (13)$$

where each column of $Y$ has zero mean. The definition of $Y$ becomes obvious by looking at $Y^T Y$:

$$Y^T Y = (\frac{1}{\sqrt{n-1}} X^T)^T (\frac{1}{\sqrt{n-1}} X^T) = \frac{1}{n-1} X^{TT} X^T = \frac{1}{n-1} X X^T = C_X, \qquad (14)$$

hence, $Y$ is an $n \times m$ and by construction $Y^T Y$ equals the covariance matrix of $X$. The PCs of $X$ are the eigenvectors of $C_X$. Applying $SVD$ to $Y$, the columns of matrix $V$ contain the eigenvectors of $Y^T Y = C_X$. Therefore, the columns of $V$ are the PCs of $X$.

$V$ spans the row space of $\equiv \frac{1}{\sqrt{n-1}} X^T$. Therefore, $V$ must also span the column space of $\frac{1}{\sqrt{n-1}} X$. We can conclude that finding the PCs amounts to finding an orthonormal basis that spans the column space of $X$. If the final goal is to find an orthonormal basis for the column space of $X$ then we can calculate it directly without constructing $Y$. By symmetry the columns of $U$ produced by the $SVD$ of $\frac{1}{\sqrt{n-1}} X$ must also be the PCs.

One benefit of $PCA$ is that we can examine the variances $C_Y$ associated with the principal components. Often one finds that large variances associated with the first $k < m$ PCs, and then a precipitous drop-off. One can conclude that most interesting dynamics occur only in the first $k$ dimensions.

Both the strength and weakness of $PCA$ is that it is a non-parametric analysis. When data are not normally distributed $PCA$ fails. In exponentially distributed data, the axes with the largest variance do not correspond to the underlying basis. There are no parameters to tweak and no coefficients to adjust based on user experience: the answer is unique and independent of the user.

This also poses a problem, if some system characteristics are not known *a-priori*, then it makes sense to incorporate these assumptions into a parametric algorithm or an algorithm with selected parameters.

This prior non-linear transformation is sometimes termed a kernel transformation and the entire parametric algorithm is called $KPCA$. This procedure is parametric because the user must incorporate prior knowledge of the structure in the selection of the kernel but it is also more optimal in the sense that the structure is more concisely described.

One might envision situations where the PCs need not be orthogonal. Only the subspace is unique because the PCs are not uniquely defined. In addition, eigenvectors beyond the rank of a matrix (i.e. $\sigma_i = 0$ for $i > rank$) can be selected almost capriciously. Nevertheless, these degrees of freedom do not influence the qualitative features of the solution nor a dimensional reduction.

For instance, if an image contains a 2-D exponentially distributed data set, then the largest variances will not correspond to the meaningful axes and *PCA* fails.

**2.3 EM algorithm for PCA**

There is a close relationship between the expectation-maximization (EM) algorithm and PCA, which leads to a faster implementation of the PPCA. The algorithm extracts a small number of eigenvectors and eigenvalues from large sets of high dimensional data. It is computationally efficient in space and time and does not require computing the sample covariance of the data.

PCA is largely used in data analysis due to its optimality in terms of mean squared error, and its linear scheme to reduce the dimensions of vectors, so that compression and decompression become simple operations to carry out given the model parameters. Notwithstanding these interesting features, PCA has some deficiencies. The other two methods for finding the PCs are impractical for high dimensional data. Difficulties can arise in both computational complexity and data scarcity when diagonalizing a covariance matrix of $n$ vectors in a $p$-dimensional space when $n$ and $p$ amount to hundreds or several thousands of elements. It is often the case that there is not enough data in high dimensions for the sample covariance to be of full rank (data scarcity). Moreover, care needs to be taken in order to use techniques such as the SVD, which do not need full rank matrices. Complexity makes the direct diagonalization of a symmetric matrix with thousands of rows tremendously expensive (it is $O(p^3)$ for $p \times p$ inputs). There are procedures such as the one proposed by Wilkinson (1965) which is $O(p^2)$ that decrease this cost when only the first most important eigenvectors and eigenvalues are necessary. The sample covariance calculation calls for $O(np^2)$ operations.

In most cases, the explicit computation of the sample covariance matrix should be avoided. Methods such as the snap-shot algorithm from Sirovich (1987), which has complexity of $O(n^3)$, take for granted that the eigenvectors sought out are linear combinations of the data points. In this section, a version of the EM algorithm from Dempster (1977) is presented for learning the PCs of a dataset. The algorithm does not require computing the sample covariance and has a complexity limited by $O(knp)$ operations, where $k$ is the number of leading eigenvectors to be learned.

Usual PCA approaches cannot handle missing values: incomplete data must either be discarded or completed via *ad-hoc* interpolation techniques. A possible and uncomplicated solution is to replace missing coordinates with the mean of the known values in the corresponding coordinate or with estimation values relying on the known values. The EM algorithm for PCA benefits from the estimation of the maximum likelihood (ML) values for missing information directly at each iteration as stated by Ghahramani & Jordan (1994).

As a final point, independently of the technique used to perform PCA, there is no accurate probability model in the input space, because the probability density is not normalized in the PS . This means that once applying PCA to some data, the only criterion on hand to verify if the new data fit well the model is the squared distance of the new data from their projections into the PS. A data point distant from the training data but close to the PS will have a high pseudo-likelihood or low error. This chapter also brings in a model called sensible PCA (SPCA) which delineates a proper covariance structure in the data space as proposed by Roweis (1998) whose main contribution was to alleviate the computational load of the other two techniques with the help of the EM algorithm.

PCA can be interpreted as a limiting case of a particular class of linear-Gaussian models (LGMs), because these models capture the covariance structure of an observed *p*-dimensional variable *y* using less than $p(p+1)/2$ free parameters when compared to the full covariance matrix calculation. LGMs do this by assuming that *y* is the result from a linear transformation of some *k*-dimensional *x* plus additive Gaussian noise. Denoting the transformation by the $p \times k$ matrix *C* and the (*p*-dimensional) noise by *v* (with covariance matrix *R*) the generative model can be written as

$$y = Cx + v x, \qquad (15)$$

where $x \sim \mathcal{N}(0, I)$ and $v \sim \mathcal{N}(0, R)$. *x* is considered independent and identically distributed (iid) according to a unit variance spherical Gaussian. Since *v* is also iid and independent of *x*, the model reduces to a single Gaussian model for *y* as follows:

$$y \sim \mathcal{N}(0, CC^T + R). \qquad (16)$$

With the purpose of saving parameters over the direct covariance representation in *p*-space, it is indispensable to select $k < p$ and to curb the covariance structure of *v* by constraining *R*. The second constraint allows the model to capture any interesting or informative projections in *x*. If *R* was not limited, then the algorithm could choose $C = 0$ and *R* would be the sample covariance of the data considering any deviation in the data as noise.

There are two central problems of interest when working with LGMs. Firstly, the compression problem asks if given fixed model parameters *C* and *R*, it is possible to gather information about the unknown *x* given a few observations *y* must be gathered. Since the data points are independent, one needs the posterior probability $P(x|y)$ given the corresponding single observation, resulting in

$$P(\mathbf{x}|\mathbf{y}) = \frac{P(\mathbf{y}|\mathbf{x})P(\mathbf{x})}{P(\mathbf{y})} = \mathcal{N}(\beta \mathbf{y}, \mathbf{I} - \beta \mathbf{C})|\mathbf{x}, \qquad (17)$$

where $\beta = \mathbf{C}^T (\mathbf{CC}^T + \mathbf{R})^{-1}$ gives the expected value $\beta y$ of the unknown and an estimate of the uncertainty in this value in the form of the covariance (I-$\beta C$). *y* from *x* can be obtained from P(*x*|*y*). Finally, the likelihood of any data point *y* comes from (16).

The second problem is called learning or parameter fitting. It seeks the matrices *C* and *R* that assign the highest likelihood to the observed data. There is a family of EM algorithms employing the inference formula above in the E-step to estimate the unknown and then choose *C* as well as *R* in the M-step, in order to maximize the expected joint likelihood of the estimated *x* and the observed *y*.

PCA is a limiting case of the LGM as the covariance of the noise *v* becomes infinitesimally small and equal in all directions, that is $\mathbf{R} = \lim_{\varepsilon \to 0} \varepsilon \mathbf{I}$. This makes the likelihood of *y* subject exclusively to the squared distance between it and its reconstruction $C_X$. The directions of the columns of $C$ which minimize this error are the *PCs*. Inference now becomes a simple least squares projection:

$$P(x|y) = \mathcal{N}(\beta y, I - \beta C)|x, \text{ with } \beta = \lim_{\varepsilon \to 0} C^T(CC^T + \varepsilon I)^{-1} \quad (18)$$

or alternatively,

$$P(x|y) = \mathcal{N}((C^T C)^{-1} C^T y, \ 0)|_x = \delta(x - (C^T C)^{-1} C^T y) \quad (19)$$

Given that the noise became insignificant, the posterior over $x$ collapses to a single point and the covariance turns out to be zero. Albeit the PCs can be computed explicitly, there is still an EM algorithm for the limiting case of zero noise. It can be easily derived from the standard algorithms (Sangers (1989), Oja (1989), Everitt (1984), Ghahramani et al. (1997)) by replacing the common E-step by the above projection as follows:

$$\text{E-step: } X = (C^T C)^{-1} C^T Y \quad (20)$$

$$\text{M-step: } C = YX^T(XX^T)^{-1}, \quad (21)$$

where $Y$ is a $p \times n$ matrix containing all the observed data and $X$ is a $k \times n$ matrix with the unknowns. The columns of $C$ span the space of the first $k$ PCs. To explicitly compute the corresponding eigenvectors and eigenvalues, the data can be projected onto this $k$-dimensional subspace to construct an ordered orthogonal covariance basis. This means that once an orientation for the PS was guessed, the presumed subspace is corrected and the data $y$ is projected onto it to give the values of $x$. Next, the values of $x$ are corrected and a subspace orientation is chosen to minimize the squared reconstruction errors of the data points.

Bear in mind that if $C$ is $p \times k$ with $p > k$ and is rank $k$ then left multiplication by $C^T(CC^T)^{-1}$, which appears not to be well defined because $(CC^T)$ is not invertible, is exactly equivalent *to* left multiplication by $(C^T C)^{-1} C^T$. This is the same as the SVD idea of defining the inverse of the diagonal singular value matrix as the inverse of an element except if it is zero when it stays zero. The perception is that even if $CC^T$ in fact is not invertible, the directions along which it is not invertible are just those that $C^T$ is about to project out.

The EM algorithm for PCA amounts to an iterative procedure for finding the $k$ leading eigenvectors without explicit computation of the sample covariance. Its complexity is limited by $O(knp)$ per iteration and so depends only linearly on both the dimensionality of the data and the number of points. Explicitly computing the sample covariance matrix result in complexities of $O(np^2)$, while other methods that form linear combinations of the data must calculate and diagonalize a matrix with all possible inner products between points and as a result have $O(n^2 p)$ complexity.

According to Roweis (1998), the standard convergence proofs for EM given by Dempster (1977) are appropriate to this algorithm as well, so a solution will always attain a local maximum of likelihood. Additionally, it is assumed that PCA learning do not have a stable maxima other than the global optimum which results in convergence to the true PS. The rate of convergence depends on the ratio of the largest eigenvalue to the second largest eigenvalue; the closer the two are in magnitude the slower the convergence will be.

In the complete data setting, the values of the projections $x$ are viewed as missing information for EM. During the E-step these values are computed by means of projecting the observed data into the current subspace. This minimizes the model error given the observed data and the model parameters. However, if some of the input points lack certain coordinate values, then those values can be easily estimated in the same fashion. The E-step can be generalized as follows:

**E-step**: For each (possibly incomplete) point $y$ find the unique pair of points $x^*$ and $y^*$ ( such that $x^*$ lies in the current PS and $y^*$ lies in the subspace defined by the known coordinates of $y$) which minimize the norm $\| Cx* - y^*\|$. Set the corresponding column of $X$ to $x^*$ and the corresponding column of $Y$ to $y^*$.

If $y$ is complete then $y^* = y$ and $x^*$ is found exactly as before. If not, then $x^*$ and $y^*$ are the solution to a least squares problem and can be found by, for instance, $QR$ factorization. Observe that this method is not restricted to missing coordinates in the data; the unknown degrees of freedom may lie in any directions in the space. This outperforms replacing each missing coordinate with the mean of known coordinates.

## 3. EM algorithm for Sensible PCA (SPCA)

If $R$ must have the form $\varepsilon I$, but do not take the limit as $\varepsilon \to 0$, then this model is called SPCA according to Roweis (1998). The columns of $C$ are still known as the PCs. From now on, the scalar value $\varepsilon$ on the diagonal of $R$ is called global noise level. It is worth noting that SPCA uses $1 + pk - k(k-1)/2$ free parameters to model the covariance. Once again, inference is done with (17) and learning by an EM algorithm. Because it has a finite noise level, SPCA defines the following model and probability distribution in the data space:

$$y \sim \mathcal{N}(0, CC^T + \varepsilon I) \qquad (22)$$

which makes possible to evaluate the actual likelihood of new test data under an SPCA model. Furthermore, this likelihood will be much lower for data far from the training set even if they are near the PS, unlike the reconstruction error from PCA. The EM algorithm for SPCA is:

$$\text{E-step: } \beta = C^T(CC^T + \varepsilon I)^{-1} \; \mu_x = \beta Y \; \Sigma_x = nI - n\beta C + \mu_x \mu_x^T \qquad (23)$$

$$\text{M-step: } C = Y\mu_x^T \Sigma^{-1} \; \varepsilon = \text{trace}\,[XX^T - C\mu_x Y^T]/n^2 \qquad (24)$$

Since $\varepsilon I$ is diagonal, the inversion in the E-step can be performed efficiently using the matrix inversion lemma:

$$(CC^T + \varepsilon I)^{-1} = (I/\varepsilon - C(I + C^TC/\varepsilon)^{-1}C^T/\varepsilon^2) \,. \qquad (25)$$

Because only the trace of the matrix in the M-step is taken, there is no need to compute the full sample covariance $XX^T$. Instead only the variance along each coordinate need to be computed. These two observations suggest that for small $k$, learning for SPCA also have complexities limited by $O(knp)$ and not worse.

## 4. EM algorithm for KPCA

Tipping & Bishop (1999) analyzed PCA from a probabilistic point of view and realized that probabilistic PCA (PPCA) is a special case of factor analysis. In addition, Rubin & Thayer (1984) developed the expectation-maximization (EM) learning algorithm for factor analysis. So, considering PPCA within the factor analysis framework, the principal components can be straightforwardly extracted by using the EM algorithm rather than performing eigenvalue decomposition. So, the computational burden on high dimensional data can be alleviated. Rosipal and Girolami (2001), transformed the EM procedure from data space to a nonlinearly related feature space. Thus, an EM approach to kernel PCA (KPCA) has arisen, which is very useful to find the nonlinear PCs. Scholkopf et al. (1998) introduced KPCA and demonstrated its value in machine learning as well as pattern recognition.

- KPCA needs to diagonalize the kernel matrix $K$, whose dimensionality $N$ is equal to the number of data points and as the data set increases, KPCA becomes less viable due to the augmenting computational complexity $O(N^3)$, which prohibits it from being used in many applications. Moreover, there is still the problem of numerical precision when diagonalizing large matrices directly according to Rosipal and Girolami, (2001). So, the EM approach to KPCA (with computational complexity $O(qN^2)$ per iteration, where $q$ is the number of extracted components) is a good remedy. Still, there exists a rotational ambiguity with the EM algorithm for PCA (and KPCA), which is unwanted from a theoretical point of view.
- Ahn & Oh (2003) have introduced a constrained EM algorithm by using a coupled latent variables model. Their proposed EM approach can directly compute the eigen-system of sample covariance matrix in data space as well as that of the kernel matrix. For the most part, when it is applied to the kernel matrix $K$, it is a dual form of the constrained EM algorithm for performing KPCA.

### 4.1 EM algorithm for any positive semi-definite matrix

Let $Y = [y_1, \ldots, y_N]$ be the matrix consisting of $N$ $p$-dimensional vectors known as observations, and $X = [x_1, \ldots, x_N]$ be the $q$-dimensional latent variables associated with the data points of $Y$. The linear model relating an observed data vector $y_n$ to a corresponding latent variable $x_n$ is given by

$$y_n = W^T x_n + \varepsilon_n, \qquad (26)$$

with $n = 1, \ldots, N$; the parametrical matrix $W \in \mathbb{R}^{p \times q}$ determines the connection between the data space, and the latent space. The $p$-dimensional noise vector $\varepsilon_n$ is normally distributed with zero mean and covariance matrix $\sigma^2 I$. Vector $x_n$ is also zero mean and normally distributed with identity covariance. By marginalizing with respect to $x_n$ and optimizing $W$ using the ML principle, Tipping & Bishop (1999) proved that the ML solution correspond to the situation when $W$ spans the PS of the observed data (PPCA model).

The EM approach to PCA is a least-squares projection, which - as said by Rosipal & Girolami (2001), Xu (1998) and Tipping & Bishop (1999) - is given by:

$$\text{E-step: } X = (W^T W)^{-1} W^T Y, \text{ and} \qquad (27)$$

$$\text{M-step: } = Y X^T (X X^T)^{-1}. \qquad (28)$$

Later, by means of a coupled latent variables model along with PPCA, Ahn & Oh (2003) introduced a constrained EM algorithm for PCA:

$$\text{E-step: } \boldsymbol{X} = \{\mathcal{L}(\boldsymbol{W}^T\boldsymbol{W})\}^{-1}\boldsymbol{W}^T\boldsymbol{Y}, \text{ and} \tag{29}$$

$$\text{M-step: } = \boldsymbol{Y}\boldsymbol{X}^T\{\mathcal{U}(\boldsymbol{X}\boldsymbol{X}^T)\}^{-1}, \tag{30}$$

where the element-wise lower operator $\mathcal{L}$ was defined such that $\mathcal{L}(a_{ij}) = a_{ij}$, for $i \geq j$ and zero otherwise, and the upper operator $\mathcal{U}$ corresponds to $\mathcal{U}(a_{ij}) = a_{ij}$, for $i \leq j$ and zero if not. Ahn & Oh (2003) verified that as the noise level became infinitesimal, $\boldsymbol{W}$ would converge to be the ML estimator $\boldsymbol{W} = \boldsymbol{U}\boldsymbol{\Lambda}^{1/2}$, where the $q$ columns of $\boldsymbol{U}$ were the eigenvectors of the sample covariance matrix $(1/N)\boldsymbol{Y}\boldsymbol{Y}^T$ and $\boldsymbol{\Lambda}$ contained the related eigenvalues arranged diagonally in descending order of magnitudes. This removes the rotational ambiguity of algorithm as stated in (27) and (28).

The EM algorithm for PCA considers the latent variables $\{x_n\}$ as missing data. The $E$-step of the LM algorithm evaluates the expectation of the corresponding complete-data log-likelihood with respect to the posterior distribution of $x_n$ given the observed $y_n$. The expectations $E(x_n|y_n)$ and $E(x_n x_n^T|y_n)$ form the basis of the E-step. In the M-step, the parameter $\boldsymbol{W}$ is updated to maximize the expected complete-data log-likelihood function, which is guaranteed to increase the likelihood of the observed samples $\{y_n\}$ as follows:

$$\boldsymbol{W} = (\sum_{n=1}^{N} y_n E(x_n|y_n)^T)(\sum_{n=1}^{N} y_n E(x_n x_n^T|y_n)). \tag{31}$$

Combining the E-step and M-stes yields

$$\boldsymbol{W} = \boldsymbol{Y}\boldsymbol{Y}^T\boldsymbol{W}(\boldsymbol{W}^T\boldsymbol{W})^{-1}((\boldsymbol{W}^T\boldsymbol{W})^{-1}\boldsymbol{W}^T\boldsymbol{Y}\boldsymbol{Y}^T\boldsymbol{W}(\boldsymbol{W}^T\boldsymbol{W})^{-1})^{-1} \tag{32}$$

as the noise level becomes infinitesimal.

Now from (32), if]et $S = \boldsymbol{Y}\boldsymbol{Y}^T$ and use the *co* upled latent variables model Ahn & Oh (2003), then the EM algorithm for PCA could be rewritten as

$$\text{E-step: } \boldsymbol{Z} = \{\mathcal{L}(\boldsymbol{W}^T\boldsymbol{W})\}^{-1}\boldsymbol{W}^T, \text{ and} \tag{33}$$

$$\text{M-step: } \boldsymbol{W} = \boldsymbol{S}\boldsymbol{Z}^T\{\mathcal{U}(\boldsymbol{Z}\boldsymbol{S}\boldsymbol{Z}^T)\}^{-1}. \tag{34}$$

The modified constrained EM algorithm comes from (33) and (34), can be further generalized to any positive semi-definite matrix $S$. In fact, by the incomplete Cholesky decomposition any positive semi-definite matrix can be factorized as

$$\boldsymbol{S} = \boldsymbol{L}\boldsymbol{L}^T, \tag{35}$$

where $\boldsymbol{L} \in \mathbb{R}^{p \times r}$ and $S$ have rank $r$. The columns of $\boldsymbol{L}$ are samples of $y_n$ and apply the model (26). The modified constrained EM algorithm comes from (33) and (34). After convergence, the normalized columns of $\boldsymbol{W}$ are the leading $q$ eigenvectors of $\boldsymbol{S}$ (please, refer to Roweis (1998) and Dempster (1977)). The related eigenvalues are the diagonal elements of $\boldsymbol{W}^T\boldsymbol{S}\boldsymbol{W}$. The computational complexity of the proposed EM algorithm is $O(qp^2)$ per iteration.

The modified constrained EM algorithm (MCEM) even though simple in the derivation, is very useful in computing not only the eigen-system of sample covariance matrix in data space, but also that of kernel-based nonlinear algorithms. For instancee, the MCEM can be applied to KPCA directly. Given the set of $N$ observations $\mathbf{Y}$, the basic idea is to first map these input data into some new feature space via a nonlinear function $\phi$, followed by standard linear PCA using the mapped samples $\phi(\mathbf{y}_i)$. Thus, KPCA turns out to compute the most important eigenvectors of the $N \times N$ kernel matrix $\mathbf{K}$ which is defined such that the elements

$$\mathbf{K}_{ij} = k(\mathbf{y}_i, \mathbf{y}_j) = (\phi(\mathbf{y}_i).\phi(\mathbf{y}_j)), \qquad (36)$$

where $k$ is the kernel function which calculates the dot product between two mapped samples $\phi(\mathbf{y}_i)$ and $\phi(\mathbf{y}_j)$. Hence, the mapping of $\phi$ does not need to be computed explicitly. If $k$ is a positive definite kernel, then there exists a mapping $\phi$ into the dot product space $\mathcal{G}$ such that (36) holds.

To compute the leading eigenvectors of $\mathbf{K}$, it is enough to replace $\mathbf{S}$ with $\mathbf{K}$ using the MCEM algorithm which can be viewed as a dual form of the constrained EM algorithm for performing KPCA. The computational complexity is $O(qN^2)$ per iteration. The projection of a test point $\mathbf{y}$, whose image is $\phi(\mathbf{y})$ onto the $q$ nonlinear principal axes is given by $\mathbf{W}^T\mathbf{K}_\mathbf{y}$, where the columns of $\mathbf{W}$ are normalized, i.e., the $i$-th column $\mathbf{w}_i$ is divided by $\sqrt{\mathbf{w}_i^T \mathbf{K} \mathbf{w}_i}$, such that the eigenvectors of the sample covariance matrix have unitary norms, and $\mathbf{K}_\mathbf{y}$ is the vector $(k(\mathbf{y}_1.\mathbf{y}), \ldots, k(\mathbf{y}_n.\mathbf{y}))^T$ (please, see Xu (1998)).

## 5. PCA in video event detection

Visual surveillance demands video sequence understanding, the detection of predefined events prone to activate an alarm, the tasks to performed, environment/scenario acquaintance and, consequently, a superior computational performance [Siebel and Maybank (2002), Fuentes and Velastin (2001)]. Nevertheless, smart SSs face the intricate task of analyzing people and their activities. A number of clues may be acquired from the investigation of people trajectories and their relations (tracking). The investigation of a single blob location or path can decide whether a person is located in an illegal area, running, jumping or hiding as in Figures (1) and (2). These data from two or more people may reveal facts regarding their interaction.

The amount of people present in a scene is called density. Events may also be classified into as position-based and dynamic-based events. SSs need to handle more than one image channel or cameras simultaneously in real time to be effectively applied to security, and this calls for a simplification of the image processing stages. A background recognition technique relying on motion detection along with a simple tracking algorithm to extract real-time blob and scene features such as blob position, blob speed and people density can help building semantic descriptions of predefined occurrences. Contrasting sequence parameters with the semantic description of the associated events related to the current scenario, the system is able to spot them and to alert the ultimate decision-maker. The case study presented here is concerned with people-oriented SSs applied to public environments.

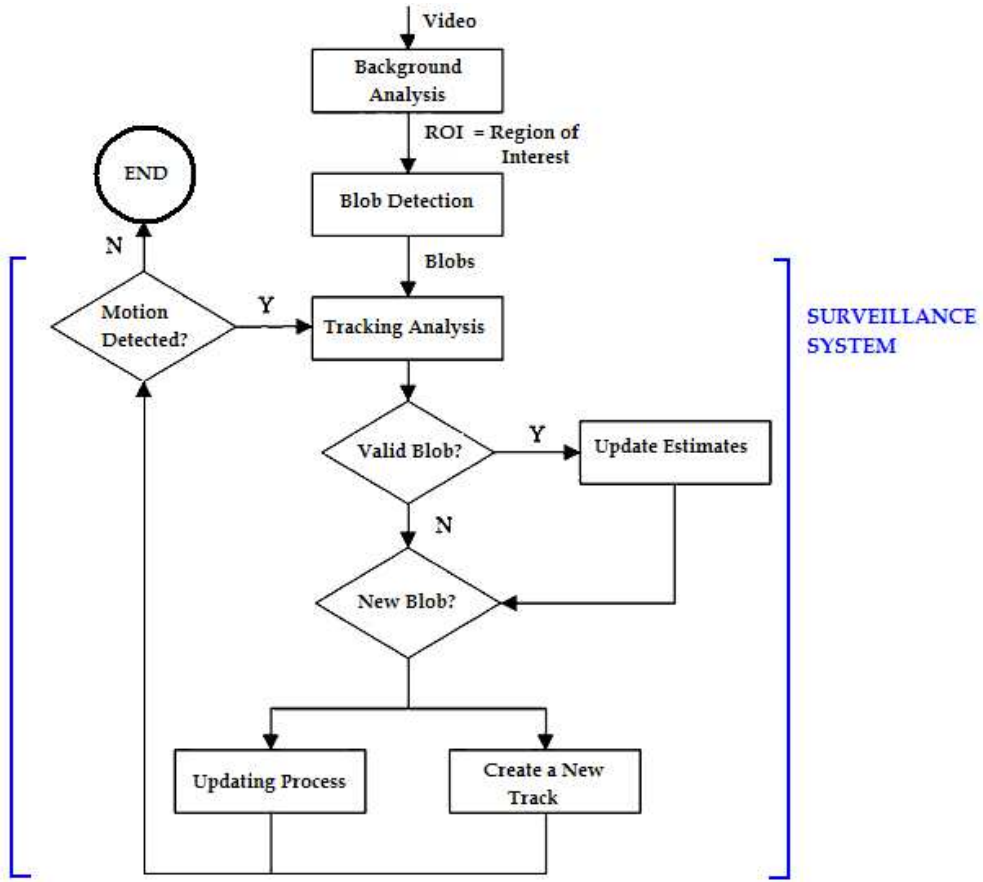

Fig. 1. Surveillance system using blob detection.

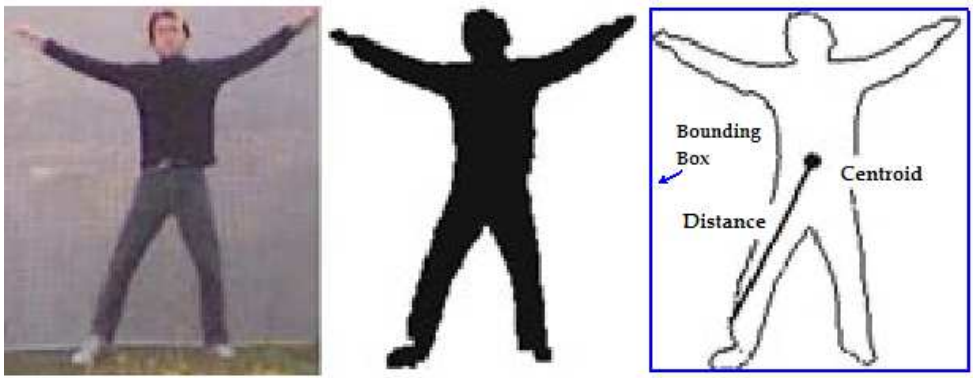

Fig. 2. Blob representation by means of a bounding box.

Background regions can be segmented out by detecting the presence or absence of motion between at least two consecutive frames as can be seen in Figure (3). An SS may hint that an event has indeed occurred because of frame-by-frame blob investigation and tracking, which offer sufficient data to analyze some prior events and the occurrence odds of others in a video sequence.

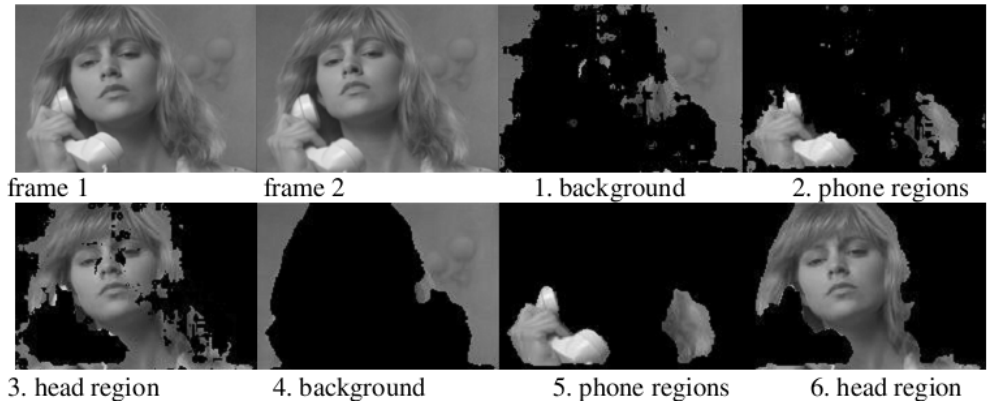

Fig. 3. Background and ROIs for the Susie sequence [Zhang et al. (2011)].

A matching procedure involving blobs from the current and preceding frames, and the overlapping of bounding boxes [Rivera et al. (2004)] can be used as evaluation criteria for ruling the interaction among people and things. This condition has proved helpful in other techniques [MacKenna et al. (2000)] and it does not necessitate the calculation of the blob position since blob motions are always considered smaller than their dimensions. Each new blob can be updated by means of the data accumulated from previous frames. If a new blob emerges, then the centroid position can be used to construct a new path. If two blobs join to form a new one, then this blob is labelled as a group and the information about combined blobs is stored for future use. This new blob group can be tracked independently. If the group splits once more, the system uses motion direction and blob characteristics to properly classify splitting blobs. Trajectories of single persons or cars can be effortlessly obtained following centroids in adjacent frames throughout a video sequence. Whenever needed, the tracked blob position can be interpolated before and after it became part of a group.

Camera networks, control rooms along with human resources and operators for surveillance have prompted lots of interest in automation of inspection tasks. These systems are also concerned with citizens' safety, people flow patterns (for counting purposes or as an aid to facilities planning), overcrowding of restricted or semi-open areas, atypical crowd movements, obstruction of exits, brawls, vandalism, falls/accidents, unattended objects, invasion of forbidden regions, and so on [Fuentes (2002)].

Blob recognition offers $2D$ information about people coordinates in a $3D$ setting. A more accurate position calls for either geometric camera calibration or stereo processing. Some examples of event detection that can be done by means of centroid position analysis [Zhang et al. (2011), Fuentes et al. (2002)] are

1. Unattended objects laying on the floor (blobs are normally smaller than people) presenting little or no motion, people falling on the ground and objects that move away from a person or reference point can be detected by means of a time analysis.
2. Suspicious activities related to hidden people may correspond to blobs disappearance along several consecutive frames.
3. Vandalism may involve isolation of one person/group present in the scene with irregular centroids motion and, possibly, changes in the background.
4. Temporal thresholding may be used to detect invasions and activate the alarms in case necessary actions must be taken.
5. Prevention of Attacks and Fights: Using people sensation of distance, and knowledge on social patterns of conduct, it is possible to establish a range of distances and profiles corresponding to different types of social interaction [MacKenna et al. (2000)]. By means of the analysis of fast centroids changes, important clues about the people present in a scene can be gathered, such as blob coincidence, merging of blobs and blob splitting to name a few.

Another way of posing the blob detection algorithm is to cluster displacement vectors and, then learn the blob centroids motions.

**5.1 Case study: Motion estimation**

Motion provides important information. Significant events, such as collision paths, object docking, sensor obstruction, object properties and occlusion can be characterized and understood with the help of the optic flow (OF). Segmenting an OF field (OFF) into coherent motion groups and estimating each underlying motion are very challenging tasks when a scene has several independently moving objects. The problem is further complicated by noise and/or data scarcity.

The main problem with motion analysis is the difficulty to get accurate motion estimates without prior motion segmentation and vice-versa. Pel-recursive (PR) schemes [Franz & Krapp (2000), Franz & Chahl (2003), Kim et al. (2005), Tekalp (1995)] can theoretically overcome some of the limitations associated with blocks by assigning a unique motion vector to each pixel.

Segmenting OF via EM algorithm for mixtures of PCs can be done successfully [Estrela & Galatsanos (2000), Tipping & Bishop (1999)] because both techniques share a close relationship. Most methods assume that there is little or no interference between the individual sample constituents or that all the constituents in the samples are known ahead of time. In real world samples, it is very unusual, if not entirely impossible, to know the entire composition of a mixture sample. Sometimes, only the quantities of a few constituents in very complex mixtures of multiple constituents are of interest [Blekas et al. (2005), Kim et al. (2005), Tipping & Bishop (1999)]. This section intends to solve OF problems by means of two different takes on PCA regression (PCR): 1) a combination of regularized least squares (RLS) and PCA ($PCR_1$); and 2) RLS followed by regularized PCA regression ($PCR_2$). Both involve simpler computational procedures than previous attempts at addressing mixtures [Blekas et al. (2005), Kim et al. (2005), Tipping & Bishop (1999), Jolliffe (2002), Wold et. al. (1983)].

### 5.1.1 Problem formulation

The displacement of every pixel in each frame forms the displacement vector field (DVF) and its estimation can be done using at least two successive frames. The goal is to find the corresponding intensity value $I_k(r)$ of the *k*-th frame at location $r = [x, y]^T$, and $d(r) = [d_x, d_y]^T$ the corresponding displacement vector (DV) at the working point *r* in the current frame by means of algorithms that minimize the DFD function in a small area containing the working point assuming constant image intensity along the motion trajectory. The perfect registration of frames will result in $I_k(r)=I_{k-1}(r-d(r))$ as seen in Figure (4). Figure (5) shows some examples of pixel neighborhoods. The DFD represents the error due to the nonlinear temporal prediction of the intensity field through the DV and is given by

$$\Delta(r;d(r))=I_k(r)-I_{k-1}(r-d(r)) .$$

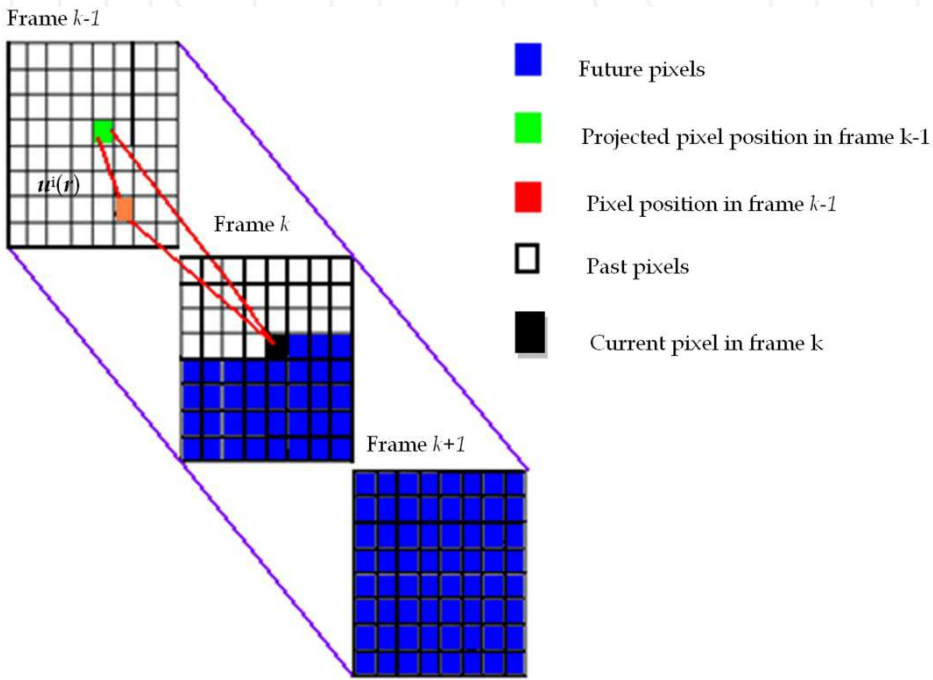

Fig. 4. Backward motion estimation problem.

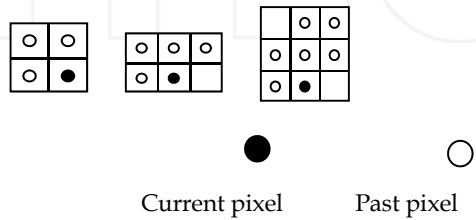

Fig. 5. Examples of causal masks.

An estimate of **d(r)**, is obtained by directly minimizing $\nabla(r,d(r))$ or by determining a linear relationship between these two variables through some model. This is accomplished by using a Taylor series expansion of $I_{k-1}(r-d(r))$ about the location $(r-d^i(r))$, where $d^i(r)$ represents a prediction of $d(r)$ in the $i$-th step. This results in $\Delta(\mathbf{r}, \mathbf{r}-\mathbf{d}^i(\mathbf{r})) = -\mathbf{u}^T \nabla I_{K-1}(\mathbf{r}-\mathbf{d}^i(\mathbf{r})) + e(\mathbf{r},d(\mathbf{r}))$, where the displacement update vector is $\mathbf{u}=[u_x, u_y]^T = d(r) - d^i(r)$, $e(r, d(r))$ stands for the truncation error resulting from higher order terms (linearization error) and $\nabla=[\partial/\partial_x, \partial/\partial_y]^T$ represents the spatial gradient operator. Considering all points in a neighborhood of pixels around $r$ gives

$$\mathbf{z} = \mathbf{Gu} + \mathbf{n},$$

where the temporal gradients $\nabla(r, r-d^i(r))$ have been stacked to form the $N\times 1$ observation vector **z** containing DFD information on all the pixels in a neighborhood, the $N\times 2$ matrix $G$ is obtained by stacking the spatial gradient operators at each observation, and the error terms have formed the $N\times 1$ noise vector $n$. The PR estimator for each pixel located at position **r** of a frame $k$ can be written as

$$d^{i+1}(r) = d^i(r) + u^i(r),$$

where $u^i(r)$ is the current motion update vector obtained through a motion estimation procedure that attempts to find $u$, $d^i(r)$ is the DV at iteration $i$ and $\mathbf{d}^{i+1}(\mathbf{r})$ is the corrected DV. The ordinary least squares (OLS) estimate of the update vector is

$$\mathbf{u}_{LS} = (\mathbf{G}^T\mathbf{G})^{-1}\mathbf{G}^T\mathbf{z},$$

which is given by the minimizer of the functional $J(u)=\|z-Gu\|^2$. The assumptions made about $n$ for least squares estimation are $E(n) = 0$, and $Var(n) = E(nn^T) = \sigma^2 I_N$, where $E(n)$ is the expected value (mean) of **n**, and $I_N$, is the identity matrix of order $N$. From now on, $G$ will be analyzed as being an $N\times p$ matrix in order to make the whole theoretical discussion easier. Since $G$ may be very often ill conditioned, the solution given by the previous expression will be usually unacceptable due to the noise amplification resulting from the calculation of the inverse matrix $\mathbf{G}^T\mathbf{G}$. In other words, the data are erroneous or noisy.

The regularized minimum norm solution also known as regularized least square (RLS) solution is given by

$$\hat{\mathbf{u}}_{RLS}(\mathbf{\Lambda}) = (\mathbf{G}^T\mathbf{G} + \mathbf{\Lambda})^{-1}\mathbf{G}^T\mathbf{z}.$$

The RLS estimate of the motion update vector can be improved by a strategy that uses local properties of the image. Each row of $G$ has entries $[g_{xi}, g_{yi}]^T$, with $i = 1, …, N$. The spatial gradients of $I_{k-1}$ are calculated through a bilinear interpolation scheme similar to what is done in [Estrela & Galatsanos (2000), Estrela & Galatsanos (1998)]. The entries $f_{k-1}(\mathbf{r})$ corresponding to a given pixel location inside a causal mask is needed to compute the spatial gradients by means of bilinear interpolation [Estrela & Galatsanos (2000), Estrela & Galatsanos (1998)] at location $\mathbf{r} = [x,y]^T$ as follows:

118    Principal Component Analysis – Engineering Applications

$$\begin{bmatrix} \theta_x \\ \theta_y \end{bmatrix} = \begin{bmatrix} x - \lfloor x \rfloor \\ y - \lfloor y \rfloor \end{bmatrix},$$

where $\lfloor x \rfloor$ is the largest integer that is smaller than or equal to $x$, the bilinear interpolated intensity $f_{k-1}(\mathbf{r})$ is specified by

$$f_{k-1}(\mathbf{r}) = \begin{bmatrix} 1-\theta_x \\ \theta_x \end{bmatrix}^T \begin{bmatrix} f_{00} & f_{10} \\ f_{01} & f_{11} \end{bmatrix} \begin{bmatrix} 1-\theta_y \\ \theta_y \end{bmatrix},$$

with. The equation evaluates the 2-nd order spatial derivatives of $f_{k-1}(\mathbf{r})$ at $r$ by means of backward differences:

$$\begin{bmatrix} g_x \\ g_y \end{bmatrix} = \begin{bmatrix} (1-\theta_y)(f_{01} - f_{00}) + (f_{11} - f_{10})\theta_y \\ (1-\theta_x)(f_{10} - f_{00}) + (f_{11} - f_{01})\theta_x \end{bmatrix}.$$

**5.2 On the use of PCA in regression**

The main idea behind the two proposed PCR procedures is the PCA of the *G* matrix [Jackson (1991), Jolliffe (2002)].

Each successive component explains portions of the variance in the total sample. PCA relates to the second statistical moment of *G*, which is proportional to *G^TG* and it partitions *G* into matrices *T* and *P* (sometimes called scores and loadings, respectively), such that:

$$\mathbf{G} = \mathbf{TP}^T.$$

**T** contains the eigenvectors of *G^TG* ordered by their eigenvalues with the largest first and in descending order. When dimensionality reduction is needed, the number of components can be chosen via examination of the eigenvalues or, for instance, considering the residual error from cross-validation [Estrela & Galatsanos (2000), Jolliffe (2002)]. The PCR motion estimation algorithms will keep the PCs and use them to group DVs inside a neighborhood. The resulting clusters will give an idea about the mixture of MVs inside a mask. The formal solution PCR$_1$ may be written as

$$\hat{\mathbf{u}}_{PCR1} = \mathbf{P}(\mathbf{T}^T\mathbf{T})^{-1}\mathbf{T}^T\mathbf{z}, \text{ and } \hat{\mathbf{u}}_{RLS}(\mathbf{\Lambda}) = (\mathbf{G}^T\mathbf{G} + \mathbf{\Lambda})^{-1}\mathbf{G}^T\mathbf{z},$$

where a regularization matrix $\Lambda$ tries to compensate for deviations from the smoothness constraint. In PCR$_1$, the scores vectors (columns in **T**) of different components are orthogonal. PCR$_1$ uses a truncated inverse where only the scores corresponding to large eigenvalues are included. The criteria for deciding when the PCR$_1$ estimator is superior to OLS estimators depend on the values of the true regression coefficients in the model. The previous solution can also be regularized:

$$\hat{\mathbf{u}}_{PCR2} = \mathbf{P}(\mathbf{T}^T\mathbf{T} + \mathbf{\Xi})^{-1}\mathbf{T}^T\mathbf{z}.$$

with Ξ standing for a regularization matrix in the PC domain. Grouping objects can be posed as a mathematical problem consisting of finding region boundaries. Sometimes the problem is such that a sample may belong to more than one class at the same time, or not belong to any class. In this method, each class is modeled by a multivariate normal in the score space from PCA. Two measures are used to determine whether a sample belongs to a specific class or not: the leverage—the Mahalanobis distance to the center of the class, the class boundary being computable as an ellipse and the norm of the residual, which must be lower than a critical value. Figure (6) shows a set of observations plotted with respect to the first two principal components (PCs). It is likely that the four clusters correspond to four different types of DVs (see ellipses). For a big neighborhood, it could happen that these vectors would not be readily distinguished using only one variable at a time, but the plot with respect to the two PCs clearly distinguishes the populations. PCR estimates are biased, but may be more accurate than OLS estimates in terms of mean square error. Nevertheless, when severe multicollinearity is suspected, it is recommended that at least one set of estimates in addition to the OLS estimates be computed since these estimates may help interpreting the data in a different way.

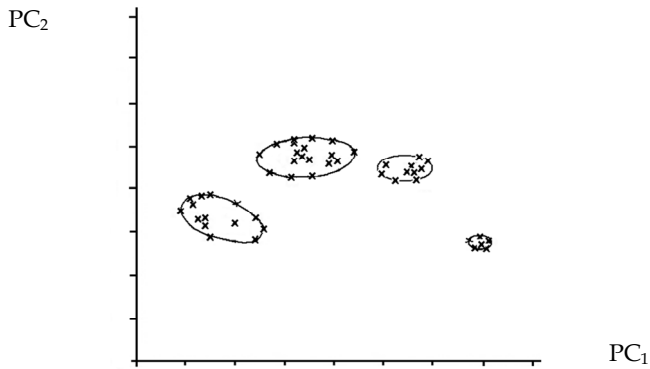

Fig. 6. An example of cluster analysis obtained by means of principal components.

When PCA reveals the instability of a particular data set, one should first consider using least squares regression on a reduced set of variables. If least squares regression is still unsatisfactory, only then should principal components be used. Besides exploring the most obvious approach, it reduces the computer load. Outliers and other observations should not be automatically removed, because they are not necessarily bad observations. As a matter of fact, they can signal some change in the scene context and if they make sense according to the above-mentioned criteria, they may be the most informative points in the data. For example, they may indicate that the data did not come from a normal population or that the model is not linear.

When cluster analysis is used for video scene dissection, the aim of a two-dimensional plot with respect to the first two PCs will almost always be to verify that a given dissection 'looks' reasonable. Hence, the diagnosis of areas containing motion discontinuities can be significantly improved. If additional knowledge on the existence of borders is used, then one's ability to predict the correct motion will increase.

PCs can be used for clustering, given the links between regression and discrimination. The fact that separation among populations may be in the directions of the last few PCs does not mean that PCs should not be used at all. In regression, their uncorrelatedness implies that each PC can be assessed independently. To classify a new observation, the least distance cluster is picked up. If a datum is not close to any of the existing groups, it may be an outlier or come from a new group about which there is currently no information. Conversely, if the classes are not well separated, some future observations may have small distances from more than one class. In such cases, it may again be undesirable to decide on a single possible class; instead, two or more groups may be listed as possible *loci* for the observation.

The average improvement in motion compensation for a sequence of *K* frames it turns out to be [Estrela & Galatsanos (2000)]:

$$\overline{IMC}(dB) = 10\log_{10}\left\{\frac{\sum_{k=2}^{K}\sum_{\mathbf{r}\in\mathbf{S}}\left[I_k(\mathbf{r}) - I_{k-1}(\mathbf{r})\right]^2}{\sum_{k=2}^{K}\sum_{\mathbf{r}\in\mathbf{S}}\left[I_k(\mathbf{r}) - I_{k-1}(\mathbf{r} - \mathbf{d}(\mathbf{r}))\right]^2}\right\}.$$

When it comes to motion estimation, one seeks algorithms that have high values of $\overline{IMC}(dB)$. A perfect registration of motion leads to $\overline{IMC}(dB) = \infty$. Figure (7) illustrates the evolution of $\overline{IMC}_k(dB)$ as a function of the frame number for two noiseless sequences: "Foreman" and "Mother and Daughter". PCR$_2$ works outperforms the other estimators due to the use of regularization in the PC domain. Figure (8) shows the DVFs for the "Rubik Cube" sequence with SNR=20 dB.

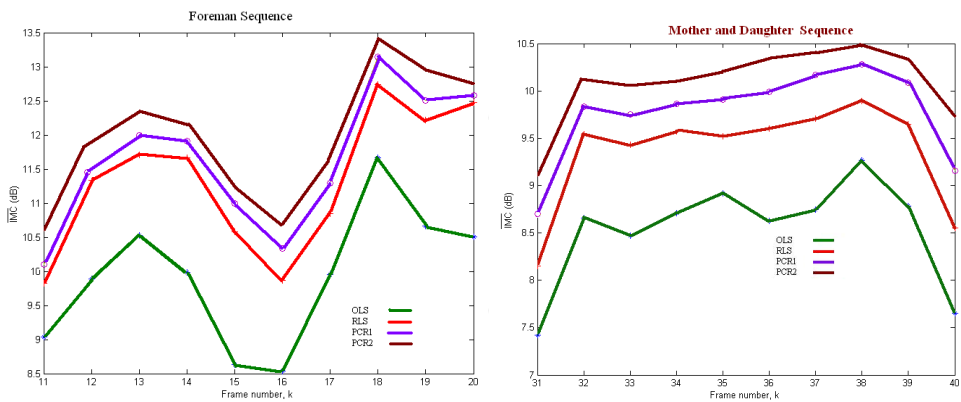

Fig. 7. Improvement in motion compensation curves for the "Foreman" and "Mother and Daughter" sequences.

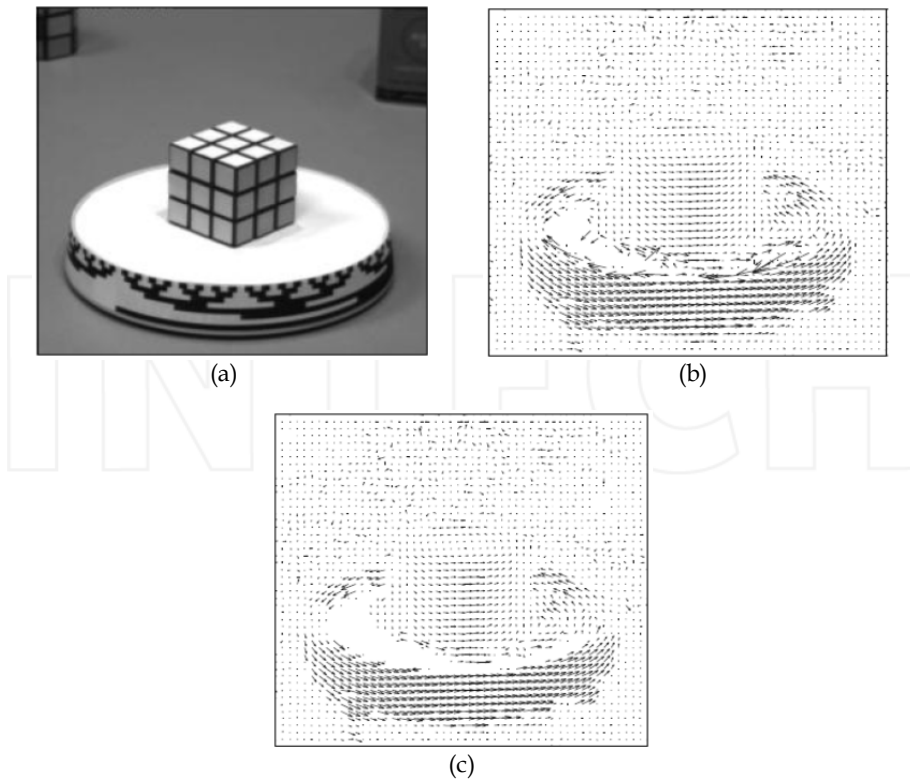

Fig. 8. Displacement field for the Rubik Cube sequence: (a) Frame of Rubik Cube Sequence; Corresponding displacement vector field for a 31×31 mask obtained by means of $PCR_1$ with SNR=20 dB; and (c) $PCR_2$, 31×31 mask with SNR=20 dB.

## 6. Final comments

The methods developed in this chapter allow simple and efficient computation of a few eigenvectors and eigenvalues when working with many data points in high dimensions. They rely on PPCA and the MCEM algorithm, which permit this calculation even in the presence of missing data.

The EM algorithms for PCA and KPCA derived above using probabilistic arguments are closely related to two well know sets of algorithms. The first are power iteration methods for solving matrix eigenvalue problems. Roughly speaking these methods iteratively update their eigenvector estimates through repeated multiplication by the matrix to be diagonalized. In the case of PCA explicitly forming the sample covariance and multiplying by it to perform such power iterations would be disastrous. However, since the sample covariance is in fact a sum of outer products of individual vectors we can multiply by it efficiently without ever computing it. In fact, the EM algorithm is exactly equivalent to performing power iterations for finding $C$ using this trick. Iterative methods for partial least squares (e.g. the NIPALS algorithm) are doing the same trick for regression. Taking the singular value decomposition (SVD) of the data matrix directly is a related way to find the PS. If the Lanczos' and Arnoldi's methods are used to compute this SVD, then the resulting iterations are similar to those of the EM algorithm. The second class of methods comprises competitive learning methods for finding the PS, such as Sanger*s* (1989) and Oja (1989) suggest. These methods enjoy the same storage and time complexities as the EM algorithm however, their update steps reduce but do not minimize the cost and so they typically need more iterations and require a learning rate parameter to be set by hand.

In this chapter, two PCR frameworks for the detection of motion fields are discussed. Both algorithms combine regression and PCA. The resulting transformed variables are uncorrelated. Unlike other works ([8, 11, 12]), the interest here is not in reducing the dimensionality of the feature space describing different types of motion inside a neighborhood surrounding a pixel. Instead, we use them in order to validate motion estimates. They can be seen as simple alternative ways of dealing with mixtures of motion displacement vectors. $PCR_1$ and PCR2 performed better than RLS estimators for noiseless and noisy images. More experiments are still needed in order to test the proposed algorithms with different types and levels of noise, so that the classification can be improved. It is also necessary to incorporate more statistical information in our models and to analyze if this knowledge will improve the outcome.

## 7. References

Ahn, J. H. & Oh, J. H. (2003). A constrained EM algorithm for principal component analysis, Neural Computation 15: 57-65.

Biemond, J., Looijenga, L., Boekee, D. E. & Plompen, R. H. J. M. (1987)A pel-recursive Wiener-based displacement estimation algorithm, Signal Proc., 13, pp. 399-412.

Blekas, K. Likas, A., Galatsanos, N.P. & Lagaris, I.E. (2005) A spatially-constrained mixture model for image segmentation, IEEE Trans. on Neural Networks,vol. 16, pp. 494-498.

Chattefuee, S. & Hadi, A.S. (2006) Regression analysis by example, John Wiley & Sons, Inc., Hoboken, New Jersey, USA.

Coelho, A., Estrela, V. V. & de Assis, J. (2009). Error concealment by means of clustered blockwise PCA. IEEE. Picture Coding Symposium. Chicago, IL, USA.

Dempster, A. P., Laird, N. M. & Rubin, D.B. (1977). Maximum likelihood from incomplete data via the EM algorithm. Proc. of the Royal Statistical Society B, 39:1–38.

do Carmo, F.P. , Estrela, V.V. & de Assis, J.T. (2009). Estimating motion with principal component regression strategies, MMSP '09, IEEE Int. Workshop on Multim. Signal Proc., Rio de Janeiro, RJ, Brazil, pp. 1–6.

Drew, M.S. & Bergner, S. (2004) Analysis of spatio-chromatic decorrelation for colour image reconstruction, 12th Color Imaging Conf.: Color, Science, Systems and Applications. Soc. for Im. Sci. & Tech. (IS&T)/Society for Inf. Display (SID) joint conference.

Estrela, V. V. & Galatsanos, N. (2000). Spatially-adaptive regularized pel-recursive motion estimation based on the EM algorithm. SPIE/IEEE Proc. of the Electronic Imaging 2000 (EI00), (pp. 372-383). San Diego, CA, USA.

Estrela, Vania V., da Silva Bassani, M.H. & de Assis, J. T. (2007) A principal component regression strategy for estimating motion, in Proc. of IASTED Int'l Conf. on Visualization, Imaging and Image Processing (VIIP2007), v.2, Mallorca, Spain, pp. 1230–1234, 2007.
Estrela, V.V. & Galatsanos, N.P. (1998) Spatially-adaptive regularized pel-recursive motion estimation based on cross-validation, Proc. of ICIP-98 (IEEE Int'l Conf. on Image Proc.), Vol. III, Chicago, IL, USA, pp. 200-203.
Everitt, B. S. (1984) An Introduction to Latent Variable Models. Chapman and Hill, London.
Franz, M.O., Chahl, J.S. & Krapp, H.G. (2004) Insect-inspired estimation of egomotion. Neural Computation 16(11), 2245-2260.
Franz, M.O. & Chahl, J.S. (2003) Linear combinations of optic flow vectors for estimating self-motion - a real-world test of a neural model, Adv. in Neural Inf. Proc. Syst., 15, pp. 1343-1350, (Eds.) Becker, S., S. Thrun and K. Obermayer, MIT Press, Cambridge, MA, USA.
Franz, M.O. & Krapp, H.G. (2000) Wide-field, motion-sensitive neurons and matched filters for optic flow fields", Biological Cybernetics, 83, pp. 185-197.
Fuentes, L.M., Velastin, S.A.: (2001) People Tracking in Surveillance Applications. $2^{nd}$ IEEE International Workshop on Performance Evaluation of Tracking and Surveillance, PETS2001.
Fuentes, L.M.: (2002) Assessment of image processing techniques as a means of improving personal security in public transport. PerSec. EPSRC Internal Report.
Fukunaga, K. (1990) Introduction to statistical pattern recognition, Computer Science and Scientific Computing. Academic Press, San Diego, 2 ed.
Galatsanos, N.P. & Katsaggelos, A.K. (1992) Methods for choosing the regularization parameter and estimating the noise variance in image restoration and their relation, IEEE Trans. Image Processing, pp. 322-336.
Jackson, J.E. (1991) A user's guide to principal components, John Wiley & Sons, Inc..
Jolliffe, I. T. (2002), Principal Component Analysis. Spinger-Verlag, 2 edition.
Kienzle, W., Schölkopf, B., Wichmann, & Franz, M.O. (2007) How to find interesting locations in video: a spatiotemporal interest point detector learned from human eye movements. Proc. of the 29th Conf. on Pattern Recognition, Heidelberg, DAGM 2007, 405-414. LNCS 4713, Springer, Berlin.
Kim, K. I., Franz, M. O. & Scholkopf, B. (2005) Iterative kernel principal component analysis for image modelling, IEEE Transactions on PAMI 27: 1351-1366.
Ma, X. , Bashir, F.I. , Khokhar, A.A. & Schonfeld, D. (2009) Event Analysis Based on Multiple Interactive Motion Trajectories, CirSysVideo(19) , No. 3, pp. 397-406.
McKenna, S., Jabri, S., Duric, Z., Rosenfeld, A., Wechsler, H.: (2000) Tracking Groups of People. Computer Vision and Image Understanding 80, 42−56
Rivera, L.A., Estrela, V.V. & Carvalho, P.C.P. (2004), Oriented Bounding Boxes Using Multiresolution Contours for Fast Interference Detection of Arbitrary Geometric Objects, Proc. of The 12-th Int'l Conf. in Central Europe on Computer Graphics, Visualization and Computer Vision (WSCG 2004), pp. 219-212.
Rosipal, R. & Girolami, M. (2001). An expectation-maximization approach to nonlinear component analysis, Neural Computation 13: 505-510.
Roweis, S.T. (1998) EM algorithms for PCA and SPCA, Advances in Neural Information Processing Systems, Vol. 10. pp. 626–632, MIT press.
Shawe-Taylor, I., & Cristianini, N. (2004) Kernel Methods for Pattern Analysis, Cambridge University Press, England.

Siebel, N.T., Maybank S. (2002) Fusion of multiple tracking algorithms for robust people tracking. Proceedings of ECCV 2002.
Sirovich, L. (1987) Turbulence and the dynamics of coherent structures, Quarterly Applied Mathematics, 45 (3):561−590.
Tekalp, A.M. (1995) Digital video processing, Prentice-Hall, New Jersey.
Tipping, M. E. and C. M. Bishop: (1999), Probabilistic principal component analysis, Journal of the Royal Statistical Society B 61(3), 611–622.
Vidal, R. Ma, Y. & Piazzi, J. (2004) A new GPCA algorithm for clustering subspaces by fitting, differentiating and dividing polynomials, IEEE Conf. on Comp. Vision and Pattern Recog., vol. I, pp. 510.517.
Wang, H., Hu, Z. & Zhao, Y. (2006) Kernel principal component analysis for large scale data set. Lecture Notes in Computer Science, 4113: 745-756.
Wang, Z., Sheikh, H. & Bovik, A. (2003), Objective video quality assessment, B. Furht, & O. Marques (Eds.), The Handbook of Video Databases: Design and Applications. CRC Press.
Yacoob, Y. & Black, M.J., (1999) Parameterized Modeling and Recognition of Activities, CVIU(73), No. 2, pp. 232-247.
Zhang, J., Shao, L. & Zhang, L., (2011) Intelligent video event analysis and understanding, Berlin-Heidelberg, Germany, Springer.
Zhang, J & Katsaggelos, A.K. (1999) Image Recovery Using the EM Algorithm, Digital Signal Processing Handbook, Ed. Vijay K. Madisetti and Douglas B. Williams, Boca Raton, CRC Press LLC.
Xu, L. (1998) Bayesian Kullback Ying-Yang dependence reduction theory, Neurocomputing 22 (1-3), 81-111.
Zheng, W., Zou, C. & Zhao, L. (2005). An improved algorithm for kernel principal component analysis, Neural Processing Letters 22: 49-56.
Zhao, R. & , Grosky, W.I. , (2002) Negotiating the semantic gap: from feature maps to semantic landscapes, PR(35) , No. 3, pp. 593-600.
Zelnik-Manor, L. , Irani, M. al, (2006) Statistical Analysis of Dynamic Actions, PAMI(28) , No. 9, pp. 1530-1535.
Ghahramani, Z. & Hinton, G. (1997) The EM algorithm for mixtures of factor analyzers, Technical Report CRG-TR-96-1, Dept. of Comp. Science, University of Toronto.
Ghahramani, Z. & Jordan, M.I. (1994), Supervised learning from incomplete data via an EM approach. In Jack D. Cowan, Gerald Tesauro, and Joshua Alspector, editors, Advances in Neural Inf. Processing Systems, volume 6, pages 120 -127. Morgan Kaufmann.
Wilkinson, J.H. (1965) The Algebraic Eigenvalue Problem. Claredon Press, Oxford, England.
Wold, S. et. alli., (1983) Pattern recognition: finding and using regularities in multivariate data, food research and data analysis, eds. H. Martens and H. Russwurm, London, Applied Science Publishers, pp. 147–188.

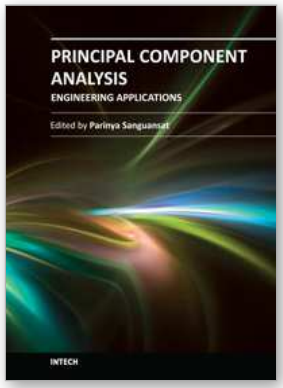

**Principal Component Analysis - Engineering Applications**
Edited by Dr. Parinya Sanguansat

ISBN 978-953-51-0182-6
Hard cover, 230 pages
**Publisher** InTech
**Published online** 07, March, 2012
**Published in print edition** March, 2012

This book is aimed at raising awareness of researchers, scientists and engineers on the benefits of Principal Component Analysis (PCA) in data analysis. In this book, the reader will find the applications of PCA in fields such as energy, multi-sensor data fusion, materials science, gas chromatographic analysis, ecology, video and image processing, agriculture, color coating, climate and automatic target recognition.

**How to reference**
In order to correctly reference this scholarly work, feel free to copy and paste the following:

Alessandra Martins Coelho and Vania Vieira Estrela (2012). EM-Based Mixture Models Applied to Video Event Detection, Principal Component Analysis - Engineering Applications, Dr. Parinya Sanguansat (Ed.), ISBN: 978-953-51-0182-6, InTech, Available from: http://www.intechopen.com/books/principal-component-analysis-engineering-applications/em-based-mixture-models-applied-to-video-event-detection